\documentclass[runningheads]{llncs}
\usepackage{graphicx}
\usepackage{booktabs} 
\usepackage{multirow}
\usepackage{bm}
\usepackage{subfigure}
\usepackage{graphicx}
\usepackage{dblfloatfix}
\usepackage{amsmath}
\usepackage{amsfonts}
\usepackage{adjustbox}
\usepackage{url}

\begin{document}
\title{Deep Attention Based Semi-Supervised 2D-Pose Estimation for Surgical Instruments}
%
%
\author{Mert Kayhan$^{1}$ \and Okan K\"op\"ukl\"u$^1$ \and Mhd Hasan Sarhan$^{1,2}$ \and Mehmet Yigitsoy$^2$ \and Abouzar Eslami$^2$ \and Gerhard Rigoll$^1$}
\institute{$^1$ Technical University of Munich, Germany\\
    $^2$ Carl Zeiss Meditec AG, Germany}

\authorrunning{M. Kayhan et al.}
\titlerunning{Semi-Supervised 2D-Pose Estimation for Surgical Instruments}
%
%
\maketitle              

\begin{abstract}

For many practical problems and applications, it is not feasible to create a vast and accurately labeled dataset, which restricts the application of deep learning in many areas. Semi-supervised learning algorithms intend to improve performance by also leveraging unlabeled data. This is very valuable for 2D-pose estimation task where data labeling requires substantial time and is subject to noise. This work aims to investigate if semi-supervised learning techniques can achieve acceptable performance level that makes using these algorithms during training justifiable. To this end, a lightweight network architecture is introduced and mean teacher, virtual adversarial training and pseudo-labeling algorithms are evaluated on 2D-pose estimation for surgical instruments. For the applicability of pseudo-labelling algorithm, we propose a novel confidence measure, total variation. Experimental results show that utilization of semi-supervised learning improves the performance on unseen geometries drastically while maintaining high accuracy for seen geometries. For RMIT benchmark, our lightweight architecture outperforms state-of-the-art with supervised learning. For Endovis benchmark, pseudo-labelling algorithm improves the supervised baseline achieving the new state-of-the-art performance.


\end{abstract}
\section{Introduction}


It has been shown that deep learning algorithms can achieve human- or super-human- level performance on variety of tasks by utilizing large amounts of labeled data. However, these achievements come at a cost: Creating these massive annotated datasets usually require a great deal of time investment, sometimes also expertise and is prone to human errors. For many practical problems and applications, it is not feasible to create such a vast and accurately labeled dataset, which restricts the application of deep learning in many areas.

A possible solution to this problem may be semi-supervised learning (SSL). Unlike supervised learning algorithms, which require all the examples to be labeled, SSL algorithms can improve performance by also leveraging unlabeled data. SSL algorithms generally enable the learning system to learn the structure of the data.  

This work investigates if the need for labels can be reduced by using semi-supervised learning in 2D-pose estimation setting. To the best of our knowledge, so far, there has not been any investigation of the usage and performance of SSL for surgical instrument tracking, where data labeling requires substantial time, and therefore, amount of unlabeled data is large compared to the labeled ones. However, this poses some fundamental challenges. In particular for 2D-pose estimation where there is no proposed method to measure the confidence of the network outputs. This is a big setback for the pseudo-labeling method where a confidence threshold is utilized to select samples where the network is certain of the answer. This study introduces total variation as a confidence measure for 2D-pose estimation task to enable the usage of pseudo-labeling.

In this work, we have applied 2D-pose estimation on surgical instruments. For this purpose, a lightweight deep attention based network architecture is proposed. On this architecture, three SSL algorithms are investigated: Mean teacher, virtual adversarial training and pseudo-labeling. Detailed experimental analysis is conducted on single-instrument Retinal Microsurgery Instrument Tracking (RMIT) dataset and multi-instrument EndoVis challenge dataset. As there is no unlabeled data for RMIT dataset, hyper parameter search is done using supervised learning. For this dataset, proposed network architecture achieves superior performance compared to state-of-the-art. For Endovis dataset, supervised learning is taken as baseline and SSL algorithms are benchmarked, where pseudo-labelling algorithm outperforms the previous state-of-the-art results.

\section{Related Work}

\noindent \textbf{Operations Requiring Surgical Tools.} Retinal microsurgery is a very challenging field for surgeons. In a typical vitreoretinal surgery, the surgeon has to manipulate retinal layers that are very delicate and less than 10 $\mu$m thick \cite{gupta1999surgical}. A surgical precision in the order of tens of microns is required for this operation. Furthermore, the resistance applied by the retinal tissue to the instruments is exceedingly small \cite{gupta1999surgical}, which limits the haptic feedback. Therefore, it is very difficult to estimate the precise location of the instruments. However, knowing where exactly the instruments are can provide vital information which can help avoid injuries inside the eye, e.g. broken blood vessel. 

Another category of surgery that can benefit from knowing exact instrument location is robotic laparoscopic surgery. Laparoscopy is a surgical procedure which examines the organs inside the abdomen to check for signs of disease. During laparoscopic surgery, small incisions are made in the wall of the abdomen and a laparoscope (a thin, lighted tube) is inserted into one of the incisions. During robotic laparoscopic surgery, surgeons receive visual information about the instruments using the cameras embedded on the robotic device \cite{sung2001robotic}. Utilizing this information, the robotic master handles are used to move the robot to the desired position. Since the surgeons are limited to the visual information collected by a rod-like instrument where left and right channels are closely embedded, estimating the depth and precise locations of instruments are very challenging. Therefore, a real-time knowledge of the instruments' position with respect to anatomical structures is a key component to improve the assistive or autonomous capabilities of surgical robots \cite{du2018articulated}.

\vspace{0.3cm}
\noindent \textbf{Approaches for Surgical Tool Pose Estimation.} Recent developments in computer vision have resulted in advanced approaches for vision-based tracking of surgical tools. The work prior to deep-learning era relies on handcrafted features, such as Haar wavelets \cite{sznitman2012unified}, gradient \cite{rieke2016real}, \cite{ye2016real} or color features \cite{zhou2014visual}, \cite{speidel2009automatic}. These approaches are not robust enough for real life scenarios due to strong illumination changes and motion blur that occur during surgeries. 

With the surge of deep learning the focus has shifted towards instrument localization and/or segmentation through CNNs. However, most of these approaches focus only on segmentation of the image, localization of keypoints on the instrument tip or bounding box detection \cite{laina2017concurrent}, \cite{sarikaya2017detection}, \cite{sahu2017addressing}, \cite{garcia2016real}, \cite{garcia2017toolnet}. The method proposed by I. Laina and N. Rieke et. al. \cite{laina2017concurrent} focuses on the interdependency between instrument segmentation and tip localization. This is the first attempt to combine these two tasks into one pipeline. By jointly optimizing for these two objectives, they improve the state of the art by a clear margin. The reported network runtime for this work is 56 ms on Nvidia TITAN X. The major shortcoming of this work is that it cannot represent the full pose of the instrument or include articulation. In response to these challenges, Du et. al. \cite{du2018articulated} provide the first work on articulated pose estimation for surgical instruments. They base their approach on the methods proposed by \cite{bulat2016human}, \cite{cao2017realtime} which consist of two stages. First, joints and joint connections are segmented, and then these are refined to come up with the final output heatmaps. These heatmaps represent the confidence of the network about the presence of a joint or joint connection at any given pixel location. Final pose of the instrument is inferred using bipartite graph matching after non-maximum suppression as post-processing step. They report a network runtime of 24 ms and post-processing runtime of 89 ms on Nvidia TITAN X GPU. Although their approach provides good generalization performance, the biggest challenge remains to be achieving real time performance while maintaining low localization error. 

\section{Methodology}

In this section, we initially give the details of the network architecture. Then we explain the proposed confidence measure, total variation, which is needed for pseudo-labeling algorithm. Finally, we mention the training details.

\subsection{Network Architecture}

For surgical tool pose estimation, a modified U-Net \cite{ronneberger2015u} architecture is used, where each joint location is found via a separate heatmap output channel. Our architecture makes use of attention mechanism intensively. Accordingly we have named our architecture DAU-Net referring to Deep Attention based U-Net. DAU-Net diverges from U-Net in the following regards: Downsampling operation is applied for 3 times, ReLU activation function is replaced with RLReLU activation \cite{xu2015empirical}, 2D attention module is added to upsampling blocks at each concatenation point, group normalization \cite{wu2018group} is used before each activation function in the main network, whereas it is omitted in the attention module. The final output maps are generated using a 1x1 convolution to scale the output channels to the number of joints and joint associations of interest. 

\begin{figure}[t!]
    \centering
    \includegraphics[width = 0.6\textwidth]{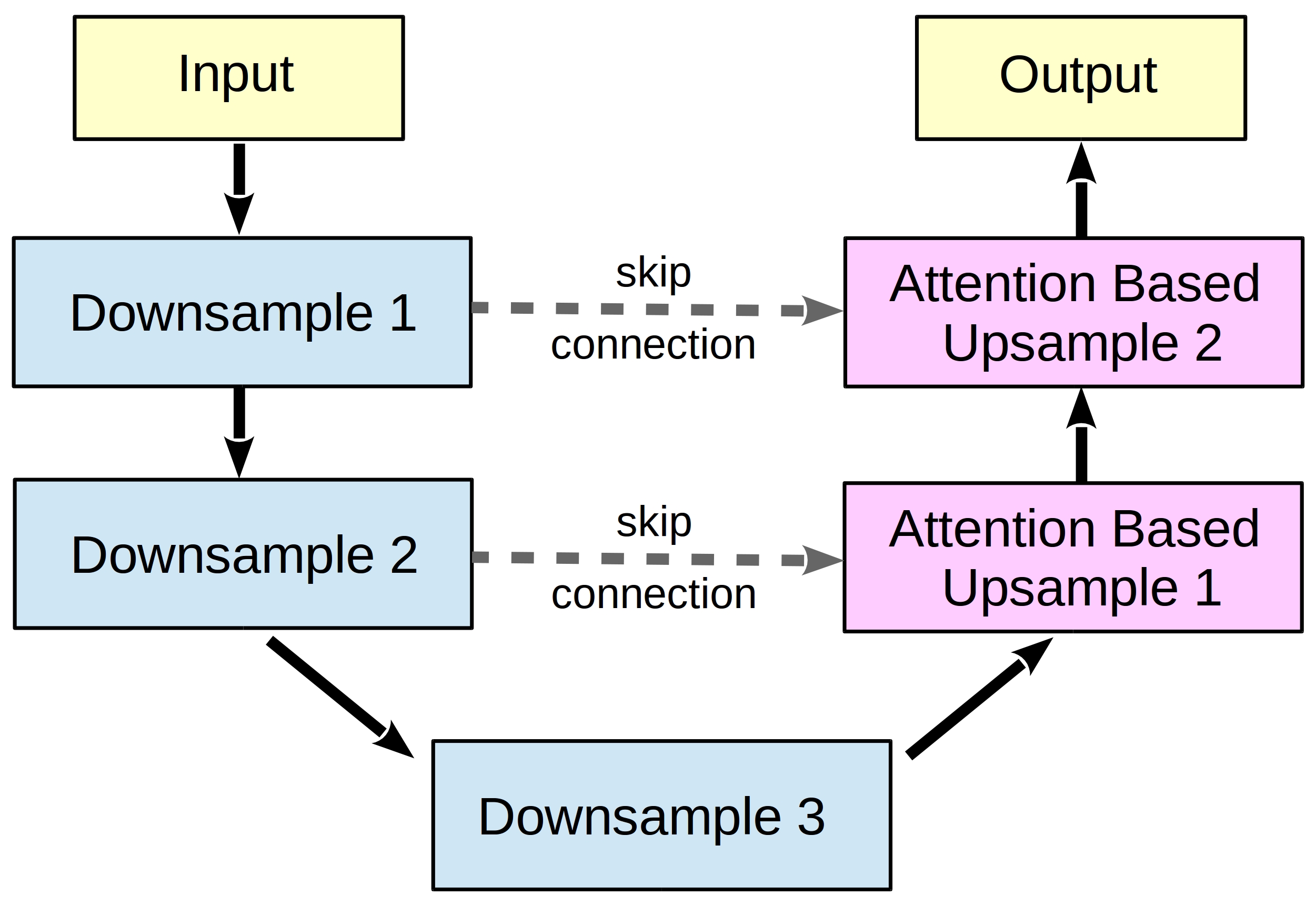}
    \caption{The modified U-Net architecture which is used in all experiments. For visual clarity, the downsample, attention based upsample and attention blocks are illustrated in Fig. \ref{fig:up_down_sample} (a), Fig. \ref{fig:up_down_sample} (b) and Fig. \ref{fig:attention_impl}, respectively.}
    \label{fig:my_unet}
    \vspace{-4mm}
\end{figure}

The final model that is used for all experiments is illustrated in Fig.~\ref{fig:my_unet}. The details of downsample and attention based upsample blocks are also illustrated in Fig. \ref{fig:up_down_sample} (a) and Fig. \ref{fig:up_down_sample} (b), respectively. Skip connections are applied from downsample block (before maxpooling) to attention based upsample blocks after deconvolution.

\subsubsection{2D Attention Mechanism for Pose Estimation}
\label{ss:attention}

Girshick et. al. \cite{girshick2015fast} have shown that by cropping relevant locations from feature maps, we can detect bounding boxes and classify the corresponding object. The biggest drawback of this method is that we need bounding box annotations to learn the correct answers.

\begin{figure}[t!]
\begin{minipage}[b]{0.480\linewidth}
\centering
    \includegraphics[height=5cm]{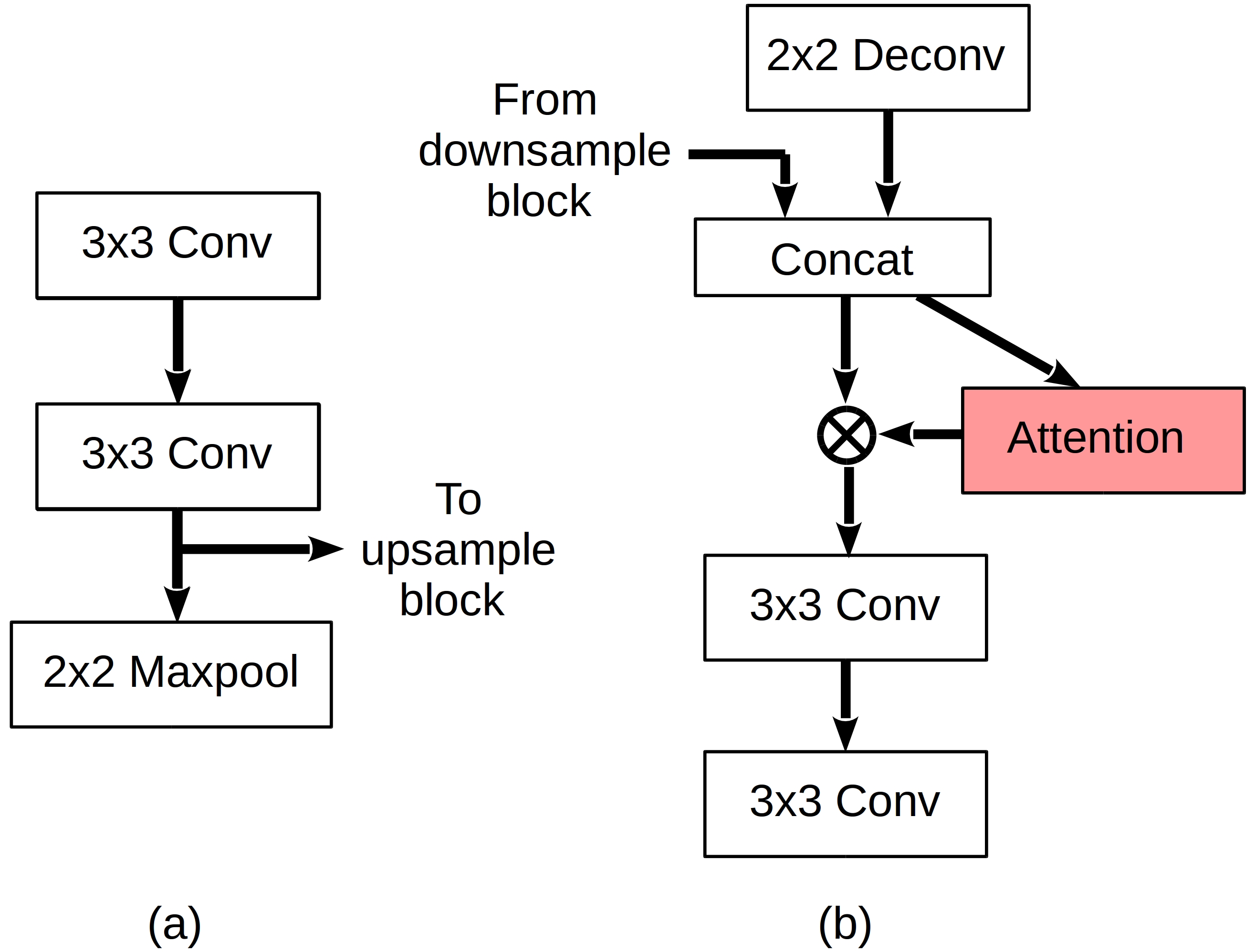}
    \caption{Architectures of downsample (a) and attention based upsample (b) blocks. Each convolution and deconvolution is followed by group normalization and RLReLU activation.  $\otimes$ stands for elementwise multiplication.} 
    \label{fig:up_down_sample}
\end{minipage}
\hfill
\begin{minipage}[b]{0.480\linewidth}
\centering
	\includegraphics[height=5cm]{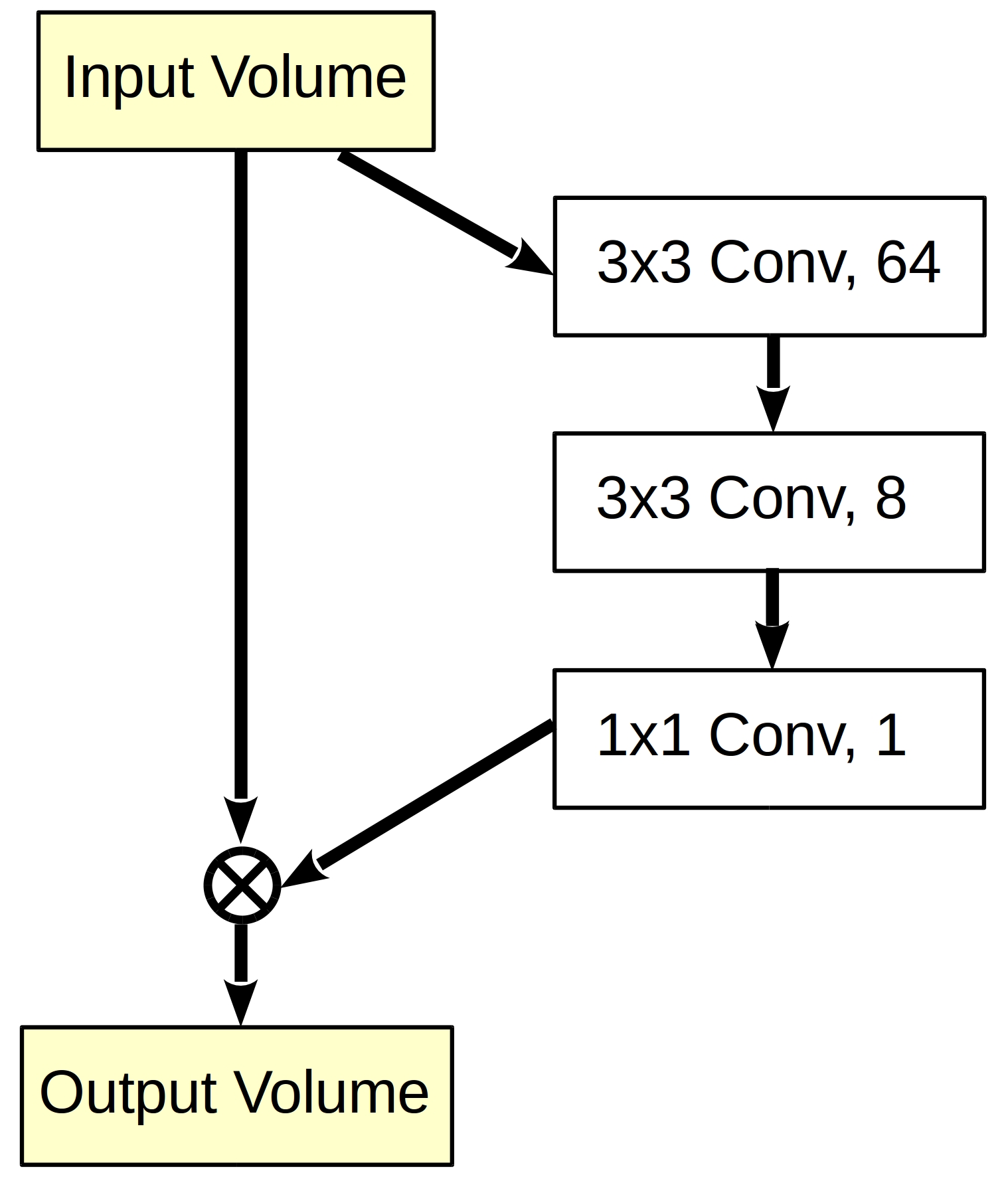}
    \caption{Attention mechanism applied in Fig. \ref{fig:up_down_sample}, where $\otimes$ refers to elementwise multiplication. Output of the attention mechanism has 1 channel with input volume size and it is broadcasted across channels during elementwise multiplication.}
    \label{fig:attention_impl}
\end{minipage}
\end{figure}



\begin{figure}[b!]
    \centering
    \includegraphics[height=3.1cm,keepaspectratio]{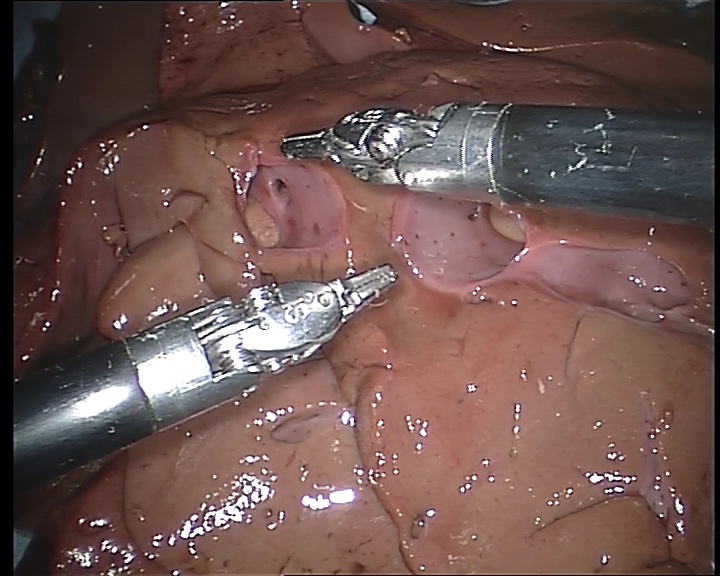} \quad \quad \quad
    \includegraphics[height=3.1cm,keepaspectratio]{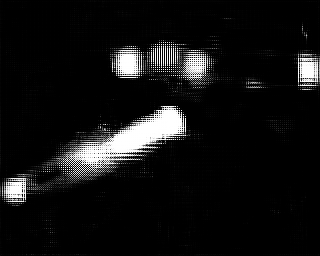}
    \caption{Visualization of the attention mechanism for multi-instrument case.}
    \label{fig:att2}
\end{figure}

To eliminate the need for bounding box annotations, 2D attention module turns on/off elements in the feature maps. The turn on/off effect is achieved by elementwise multiplication after sigmoid activation. In other words, for each element in the feature map, the attention mechanism tries to decide if this element contains information about the joints and/or connections between joints. This leads to a drastic reduction in search space for the network because only relevant elements are propagated further. The applied attention architecture is depicted in Fig. \ref{fig:attention_impl}. A visualization of the learned attention maps and the corresponding images can be seen in Fig. \ref{fig:att2}. As can be seen, the attention mechanism successfully concentrates on the important parts of the input image.

\subsection{Post-processing}
\label{ss:2d_post}

For single instrument localization only the joint probability maps are predicted, whereas for multi-instrument \mbox{localization} the connection probability maps are predicted as well. The following procedure is used to retrieve the final joint locations. 

For single instrument detection, Gaussian filter is applied to the output, and then, for each channel of the output the pixel location that contains the maximum value is found. 

For multiple instrument detection, Gaussian filter is applied to the joint probability maps which is followed by thresholded non-maximum suppression to retrieve the joint candidates. Then, if total variation measure of the output maps are below a certain threshold high-boost filter is applied to the connection probability maps. Finally, line integral \cite{cao2017realtime}, \cite{du2018articulated} is utilized to find joint pairs and the instrument is parsed.




\subsection{Total Variation as a Confidence Measure for Pose Estimation}
\label{ss:tv}

In mathematics, total variation is a measure that describes the local and global structure of functions \cite{scherzer2009variational}. Furthermore, in the context of image processing, it is often assumed that signals with high total variation have excessive detail. Following this notion, this study proposes total variation of probability maps as a way of assessing the confidence of the inferred pose estimates.  

Formally, the $\textit{anisotropic}$ version of total variation is shown as
\begin{equation*}
    V(y) = \sum_{c}\sum_{ij} | y_{i+1, j, c} - y_{i, j, c} | + | y_{i, j+1, c} - y_{i, j, c} |
    \vspace{-3mm}
\end{equation*}

$\newline$
for multi-channel images \cite{scherzer2009variational}. As can be seen in the above given formulation, total variation is the sum of the local discrete gradients in x and y direction. In other words, images with high total variation have large value differences between neighboring pixels. This is often assumed to be noise and irrelevant information, and therefore, $\textit{total variation denoising}$ \cite{vogel1996iterative} has been proposed to eliminate the noise from the images. However, in the context of CNN based 2D pose estimation, the global structure of the output maps match the instrument location because during training MSE objective is minimized. Exploiting this information, total variation of output maps can be used to evaluate the local properties of the output maps. As it can be seen in the autoencoder literature \cite{tschannen2018recent}, two images may have low MSE but look quite different because MSE does not necessarily address the sharpness of the image. In this study, by using total variation, the sharpness of the output maps is evaluated. In other words, if an output map has low total variation, this translates to a flat output distribution which represents a low confidence prediction. Thus, total variance measure can be used as a post-processing step to evaluate the quality of predictions and if necessary, enable a decision mechanism which can be used to evaluate the need for further processing. Furthermore, this measure complements the pseudo-labeling method for pose estimation because this method requires a confidence threshold to be used effectively.


\subsection{Training Details}
\label{ss:2D_training}

\noindent \textbf{Learning:} Throughout the training Adam solver is used with default parameters \cite{kingma2014adam}. The training lasted 50k iterations. Following Du et. al. \cite{du2018articulated}, input resolution is set to $288 \times 384$ pixels and $256 \times 320$ pixels for RMIT and EndoVis datasets respectively. DAU-Net kernels are initialized from a truncated Gaussian distribution and kernels in attention module is initialized using Xavier initialization. Target labels are created by heatmaps, where each joint annotation corresponds to a 2D Gaussian density map centred at the labelled point location and the annotation for joint association corresponds to a Gaussian distribution along the joint pair center line. Following Du et. al. \cite{du2018articulated}, the standard deviation of 20 pixels is used for Gaussian distributions.

\vspace{0.15cm}
\noindent \textbf{Regularization:} The network is regularized using dropout \cite{srivastava2014dropout} with dropping rate of 30\% and \%10 for RMIT and EndoVis, respectively. Also, noisy labels are applied by sampling a random variable uniformly between -0.01 and 0.01, and adding to each pixel of the target heatmap.

\vspace{0.15cm}
\noindent \textbf{Augmentation:} Since both datasets contain very limited data, heavy data augmentation is used to avoid overfitting. For RMIT dataset, random flipping, random translation (5 px), random rotation (10 degrees), Gaussian noise, random brightness, random contrast, random saturation, histogram equalization, random blurring, pepper noise, salt noise, speckle noise
and random erasing \cite{zhong2017random} are used. For EndoVis dataset,  random flipping, random translation (5 px), random rotation (20 degrees), random swapping are used.

\begin{figure}[t!]
    \centering
    \includegraphics[height=3.1cm,keepaspectratio]{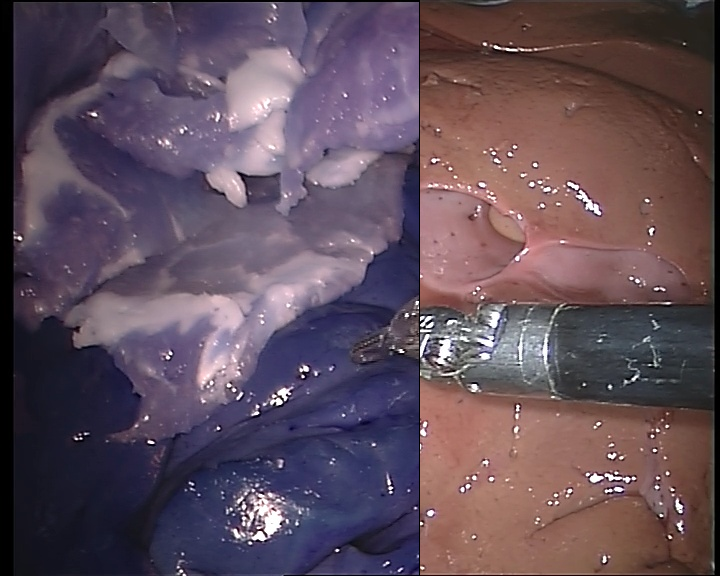} \quad \quad \quad
    \includegraphics[height=3.1cm,keepaspectratio]{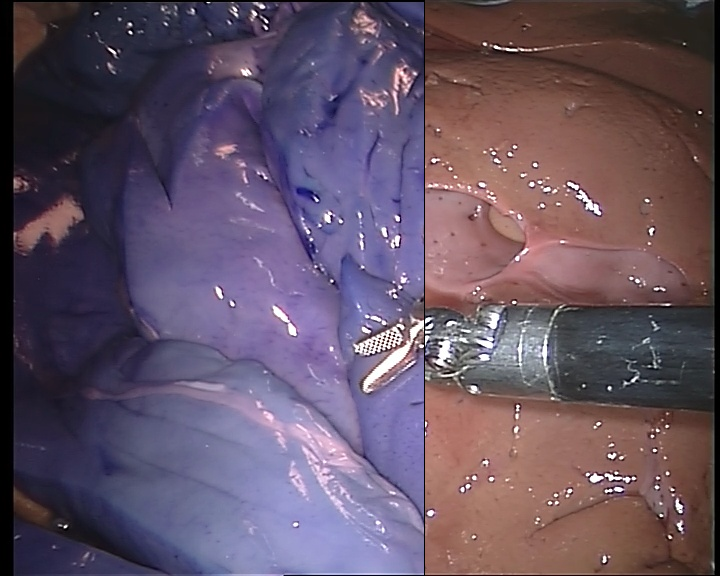}
    \caption{The resulting frames from random swapping is depicted in the figure. To make it visually more comprehensible, parts that come from different images are visualized using BGR and RGB color formats respectively.}
    \label{fig:swap}
    \vspace{-4mm}
\end{figure}

\textit{Random Swapping Data Augmentation}: EndoVis dataset contains very limited annotated samples. Furthermore, a large fraction of these samples consist of frames where only a single instrument is visible. This makes it very difficult to learn models with good generalization across single and multi-instrument cases. To deal with this issue, $\textit{Random}$ $\textit{Swapping}$ is introduced as a data augmentation strategy.

Inspired from \cite{devries2017improved}, \cite{dvornik2018modeling}, $\textit{Random}$ $\textit{Swapping}$ is a method that uses keypoint annotation to generate semantically meaningful mixtures of images and simulate occlusion. In this study, the clasper annotations are used to split the frames into 2 parts. Afterwards another image is sampled from the training set and again a split is formed depending on the clasper annotation. Finally, two cropped parts from these two training images are fused together. If the sum of the crop sizes do not correspond to the original frame size, the final image is either zero-padded from the middle or cropped from the edges. The same operations are performed on the target heatmaps as well to generate labels for training. An illustration of the output images can be seen in Fig.~\ref{fig:swap}.

\vspace{0.15cm}
\noindent \textbf{Implementation:} The whole training setup and network is implemented using Tensorflow. For reproduciblity of results, we make our code publicly available  \footnote{https://github.com/mertkayhan/SSL-2D-Pose}.

\section{Experiments}
\label{sec:exp}

In this section, we share the obtained results from our experiments on two publicly available datasets: RMIT \footnote{https://sites.google.com/site/sznitr/code-and-datasets} and Endovis \footnote{https://endovissub-instrument.grand-challenge.org}. First, RMIT dataset is used to develop the \textit{deep attention U-Net} (DAU-Net) model. Since RMIT dataset does not contain any unlabeled samples we do not investigate semi-supervised learning on this dataset. Next, EndoVis dataset is utilized to evaluate the performance of the developed network architecture. Furthermore, the unlabeled training data is used to evaluate the effectiveness of mean teacher, virtual adversarial learning and pseudo labeling algorithms.  

\subsection{Datasets}

\begin{figure}[t!]%
    \centering
    \subfigure[]{
    \includegraphics[height=3.1cm]{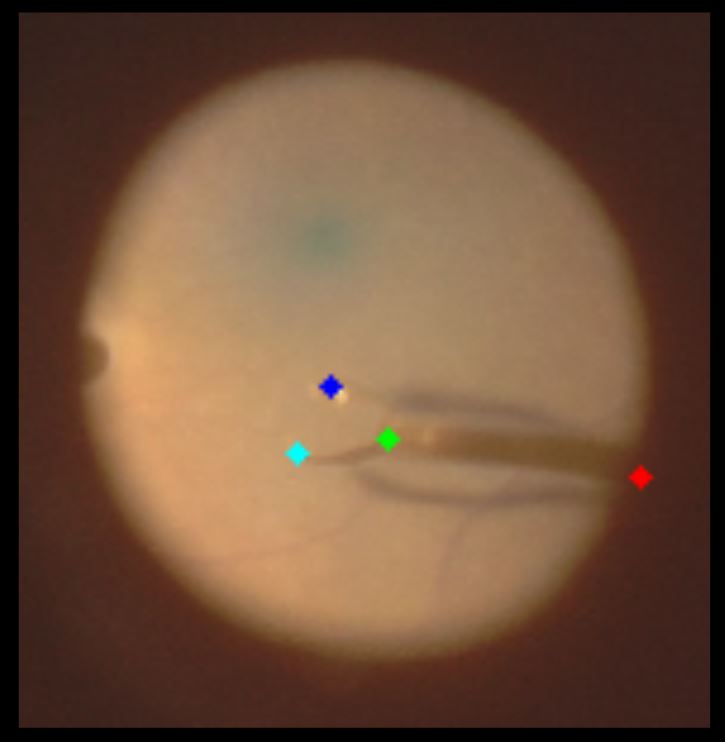}
    \label{fig:RMIT_label}}%
    \quad \quad \quad
    \subfigure[]{
    \includegraphics[height=3.1cm]{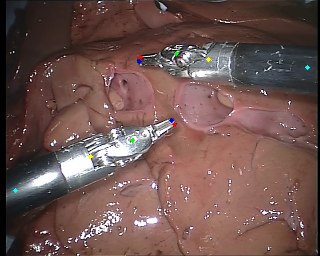}
    \label{fig:Endovis_label}}%
    \caption{Example training labels for RMIT (a) and Endovis (b) datasets. For RMIT dataset, the tips (cyan, blue), shaft (green) and end (red) of the instrument are annotated. For Endovis dataset, the claspers (red, blue), head (green), shaft (yellow) and end (cyan) joints are annotated.}
    \label{fig:labels}
\end{figure}

\noindent \textbf{RMIT Dataset:} Retinal Microsurgery Instrument Tracking (RMIT) dataset consists of three surgical sequences which are recorded during $\textit{in vivo}$ retinal microsurgery where only a single instrument is visible during recording. The original frames extracted from the videos have a resolution of 640 x 480 pixels. Following Du et. al. \cite{du2018articulated}, the dataset was split into training and test datasets where the training set consists of the first halves of each sequence and rest of the data was used for testing. A detailed distribution of the data can be seen in Table \ref{table: dataset_numbers}. For most of the frames 4 keypoints (tip1 - tip2 - shaft - end) are annotated. An example annotation can be seen in Fig.~\ref{fig:RMIT_label}. 

\vspace{0.15cm}
\noindent \textbf{Endovis Dataset:} EndoVis challenge dataset is a multi-instrument dataset that contains 6 video sequences from endoscopic surgeries where in fraction of the sequences, 2 instruments are present in the frame. The training set consists of four 45 seconds $\textit{ex vivo}$ video sequences of surgeries whereas the test set consists of four 15 seconds video sequences which are complementary to the training set as well as two additional 1 minute recorded interventions. A detailed distribution of the data can be seen in Table \ref{table: dataset_numbers}. The frame resolution for each of the videos is 720 x 576 pixels. Since the sparse annotations proposed by Du et. al. \cite{du2018articulated} are used, as done by Du et. al., the entire training set is used for training which differs from the leave-one-surgery-out training strategy requirement of the original challenge. For semi-supervised learning, the unlabeled training data is used as well. Du et. al. construct a high quality multi-joint annotation which consists of Left Clasper, Right Clasper, Head, Shaft and End joint positions. An example annotation can be seen in Fig.~\ref{fig:Endovis_label}.

\begin{table}[t!]
\centering
\small
\begin{tabular}{lcccc}
\specialrule{.15em}{.3em}{.3em}
  & \multicolumn{2}{c}{  \phantom{aaa} \textbf{EndoVis Dataset}  \phantom{aaa}} & \multicolumn{2}{c}{\phantom{aaa} \textbf{RMIT Dataset} \phantom{aaa}} \\ \cmidrule(lr){2-3} \cmidrule(lr){4-5}
  & \textbf{Training} & \textbf{Testing} & \textbf{Training} & \textbf{Testing} \\
 \specialrule{.15em}{.3em}{.3em}
 \textbf{Seq 1} \hspace{0.2cm}    & 210 / 1107    & 80 / 370    & 201 / 201     & 201 / 201     \\ 
 \textbf{Seq 2} \hspace{0.2cm}    & 240 / 1125    & 76 / 375    & 111 / 111     & 111 / 111     \\ 
 \textbf{Seq 3} \hspace{0.2cm}    & 252 / 1124    & 76 / 375    & 265 / 271     & 266 / 276     \\ 
 \textbf{Seq 4} \hspace{0.2cm}    & 238 / 1123    & 76 / 375    & ---           & ---           \\ 
 \textbf{Seq 5} \hspace{0.2cm}    & ---           & 301 / 1500  & ---           & ---           \\ 
 \textbf{Seq 6} \hspace{0.2cm}    & ---           & 301 / 1500  & ---           & ---           \\ 
 \textbf{Total} \hspace{0.2cm}    & 940 / 4479    & 910 / 4495  & 577 / 583     & 578 / 588     \\ 
 \specialrule{.15em}{.3em}{.3em}
\end{tabular}
\caption{The distribution of the data across different sequences for RMIT and Endovis datasets. Each row contains number of labeled images / number of total images for corresponding sequence. It should be noted that Sequence 5 and 6 are only present in the test set for Endovis Dataset.}
\label{table: dataset_numbers}
\end{table}

\subsection{Results Using RMIT Dataset}

For the experiments shown in Table \ref{table:RMIT1}, the network is trained on the groundtruth bounding boxes to enable faster experimentation and simulate an object detection based localization system. Bounding boxes are extracted using 3-point annotation as shown in Fig.~\ref{fig:labels}(a) excluding end point. A resolution of 128x128 is used for all experiments. In Table~\ref{table:RMIT1}, the pixel error rate corresponds to mean absolute error since the groundtruth bounding boxes are used for training which eliminates the possibility of false detection.

\begin{table}[t!]
\centering
\small
\begin{tabular}{ |l|c|c|c|c| }
 \hline
 \multicolumn{5}{|c|}{Pixel Error Rates (MAE)} \\
 \hline
  & Shaft & Tip 1 & Tip 2 & Avr \\
 \hline
 U-Net + augmentation	   & 5.9 & 7.12 & 5.57 & 6.2 \\ \hline
 U-Net + augmentation + attention  & 3.79 & 6.67 & 4.72 & 5.06 \\ \hline
 U-Net + heavy augmentation + attention & 3.73 & 5.98 & \textbf{4.12 } & 4.61 \\ \hline
 \begin{tabular}[c]{@{}l@{}}U-Net + heavy augmentation + attention \\ + L2 regularization\end{tabular} & 4.72 & 6.63 & 4.98 & 5.44 \\ \hline
 \begin{tabular}[c]{@{}l@{}}U-Net + heavy augmentation + attention \\ + dropout\end{tabular} & 3.81 & 5.71 & 4.4 & 4.64 \\ \hline
 \begin{tabular}[c]{@{}l@{}}U-Net + heavy augmentation + attention \\ + dropout + noisy labels\end{tabular} & 3.19 & 5.59 & 4.23 & 4.34 \\ \hline
 \begin{tabular}[c]{@{}l@{}}U-Net + heavy augmentation + attention \\ + dropout + noisy labels + lrelu\end{tabular} & 3.1 & 6.59 & 4.49 & 4.73 \\ \hline
 \begin{tabular}[c]{@{}l@{}}U-Net + heavy augmentation + attention \\ + dropout + noisy labels + rlrelu\end{tabular}	& \textbf{2.8 } & \textbf{4.82 } & 4.17 & \textbf{3.93 } \\
 \hline
\end{tabular}
\vspace{0.2cm}
\caption{A compact summary of the results obtained by varying one component at a time to find the right architecture and training pipeline for RMIT dataset. Augmentation refers to only geometric transformations while heavy augmentation also includes color space transformations.}
\label{table:RMIT1}
\end{table}

First, a vanilla U-Net is trained using only the geometric augmentations. It is observed that the network produces very coarse output maps which lead to high pixel error. In response to this observation, attention mechanism is introduced to help the network to concentrate on the important parts of the image. Following this modification, it is observed that the network trains a lot faster and produces more finegrained outputs. However, it is also observed that the network is highly prone to overfitting, and therefore, more data augmentation is introduced to deal with this. Even with the additional augmentation, it is observed that the network overfits, thus, regularization is added in the following experiments. It can be seen that dropout yields superior performance compared to L2 regularization, and therefore, dropout with 30\% drop rate is used in the following experiments. Because increasing the drop rate does not increase the generalization performance, more creative ways of regularizing the network are investigated. It is observed that the combination of heavy data augmentation and noise injection to the labels simulates new data points more convincingly and leads to a better generalization performance. The structure of the injected noise is described in Subsection \ref{ss:2D_training}. Finally, an investigation over the activation functions is conducted to see if generalization performance can be improved. It is observed that RLReLU activation function \cite{xu2015empirical} improves the generalization performance furthermore. All in all, one can see that by increasing the input and network level stochasticity (random data augmentation, dropout and RLReLU), the generalization performance is improved drastically. This model is used throughout this study and represented by DAU-Net-$<$$\textit{base\_feature\_maps}$$>$-$<$$\textit{depth}$$>$ which corresponds to DAU-Net-32-3 in this case.


In order to make a fair comparison with state-of-the-art, the proposed system is scaled up and trained on 4-point annotations in end to end manner. Except increasing the number of trainable parameters, no other modifications are made and same input resolution as Du et. al. \cite{du2018articulated} is used. As can be seen in Table \ref{table:comp_4_point}, DAU-Net-64-3 improves the state of the art while using fewer parameters ($\sim$2.1M) for a detection threshold of 15 pixels on the original frame. It should be noted that in Table~\ref{table:comp_4_point}, the pixel error does not correspond to mean absolute error but to root mean squared error computed for the detected joints.

\begin{table}[t!]
\centering
\small
\begin{tabular}{|l|c|c|}
 \hline
 \multicolumn{3}{|c|}{Precision / Recall / Pixel error (RMSE)} \\
 \hline
  & DAU-Net-64-3 & Du et. al. \cite{du2018articulated}  \\
 \hline
    Tip1 & 96 / 96 / 4.44 & \textbf{99.13 / 99.13 / 5.26 } \\ 
    Tip2 & \textbf{98.3 / 98.3 / 5.13 } & 97.58 / 97.58 / 4.61 \\  
    Shaft & \textbf{99.5 / 99.5 / 4.01 } & 94.12 / 94.12 / 4.93 \\ 
    End & \textbf{92.4 / 92.4 / 5.68 } & 86.51 / 86.51 / 4.68 \\  \hline
    Avr & \textbf{96.6 / 96.6 / 4.82 } & 94.3 / 94.3 / 4.87 \\ 
 \hline
\end{tabular}
\vspace{0.2cm}
\caption{Comparison of DAU-Net-64-3 with the state of the art for 4-point annotation and end-to-end training on RMIT dataset.}
\label{table:comp_4_point}
\end{table}

\subsection{Results Using Endovis Dataset}
\label{ss:endovis_results}

\begin{table}[b!]
\centering
\small
\begin{tabular}{ |p{3cm}| p{2cm} |  }
 \hline
 \multicolumn{2}{|c|}{Experiments (Test set loss (MSE))} \\
 \hline
  & Avr. Loss \\
 \hline
30\% dropout & 0.002337 \\
Dilated Conv & 0.003029 \\
10\% dropout  & 0.002357 \\
Random Swap & \textbf{0.002288} \\
Elastic Disp. & 0.002302 \\ \hline
\end{tabular}
\vspace{0.2cm}
\caption{A compact summary of the results obtained by varying one component at a time to find the right amount of regularization and data augmentation strategies for EndoVis dataset.}
\label{table:Endovis1}
\end{table}

$\newline$
For all the experiments given in Table \ref{table:Endovis1}, $\textit{DAU-Net-64-3}$ is used because it was shown to deliver very accurate pose estimates for single instrument cases. The main idea of these experiments is to test the performance on multi-instrument cases and measure the effectiveness of semi-supervised learning in 2D-pose estimation setting. Since finding the exact poses of multiple instruments require a post-processing procedure based on thresholded non-maximum suppression and graph matching, test set loss is compared to find models with better performance.

First, the network is trained with the exact setup from the previous section. However, it is observed that the generalization performance is not very good. At the beginning it is speculated that this is caused by the larger receptive field requirement for the EndoVis dataset. Therefore, dilated convolutions with dilation rate 2 are introduced. As can be seen in the Table~\ref{table:Endovis1}, this does not improve the performance. After analysing the output maps, it is observed that network produces flat outputs to minimize MSE which is interpreted as underfitting. In response to that, dropout rate is reduced to 10\%. Furthermore, the color space augmentations are removed because in EndoVis the lighting does not vary between sequences. Next, random swapping data augmentation is introduced to generate more data. Introduction of random swapping reduced the test set error below 0.0023. Afterwards, to see the effectiveness of random swap, it is removed from the augmentation pipeline and elastic displacement is introduced. However, this model performs slightly worse.

\begin{table}[t!]
\centering
\small
\begin{tabular}{|l|c|c|c|c|c|c|c|c|}
 \hline
 	\multicolumn{1}{|c|}{\multirow{3}{*}{Keypoint}} & \multicolumn{4}{|c|}{Sequence 1-4 (Seen instruments)} & \multicolumn{4}{|c|}{Sequence 5-6 (Unseen instruments)} \\
 \cline{2-9}  
  & Precision &  Recall & F1-score & \begin{tabular}[c]{@{}c@{}}Pixel error \\ (RMSE)\end{tabular} & Precision &  Recall & F1-score & \begin{tabular}[c]{@{}c@{}}Pixel error \\ (RMSE)\end{tabular}   \\
 \hline
    Left Clasper     & 95.6 & 100 & 97.8 & 4.44     & 58.0 & 83.5 & 68.5 & 8.13 \\ 
    Right Clasper    & 99.7 & 100 & 99.9 & 2.83     & 90.7 & 63.5 & 74.7 & 5.85 \\  
    Head             & 99.7 & 100 & 99.9 & 4.23     & 95.1 & 65.1 & 77.3 & 4.92 \\ 
    Shaft            & 100  & 100 & 100  & 2.86     & 99.4 & 66.1 & 79.4 & 8.11 \\  
    End              & 100  & 100 & 100  & 5.93     & 91.9 & 56.1 & 69.7 & 7.13 \\ \hline
    Avr              & 99.1 & 100 & 99.5 & 4.06     & 87.0 & 68.7 & 73.9 & 6.83 \\ 
 \hline
\end{tabular}
\vspace{0.2cm}
\caption{Performance of the supervised baseline on the seen and unseen instruments after \mbox{post-processing}.}
\label{table:endovis2}
\end{table}


Table \ref{table:endovis2} shows the precision, recall, f1-scores and the RMSE of the network for seen and unseen instruments. As can be seen, the network delivers very accurate pose estimates for seen instruments compared to unseen instruments since the network has difficulty extrapolating to an unknown geometry. Except for the left clasper, it can be seen that the detected joints are mostly within the 20 pixel threshold, whereas for left clasper, detections are not very accurate. After analysing the output maps, it is seen that the network produces low confidence predictions which get thresholded away. It is speculated that this is the main reason for the low recall for most of the joints. To counter that, our proposed total variation confidence measure is utilized. More information about this method can be found in Section \ref{ss:tv}. Using the steps given in Section \ref{ss:2d_post}, an improvement from 73.9 to 76.1 in average f1-score is observed. A detailed report of the results with this new post-processing pipeline can be seen in Table \ref{table:endovis4}.

\begin{table}[t!]
\centering
\small
\begin{tabular}{|l|c|c|c|c|}
 \hline
 \multicolumn{5}{|c|}{Sequence 5-6 (Unseen instruments)} \\
 \hline
  & Precision &  Recall & F1-score & \begin{tabular}[c]{@{}l@{}}Pixel error \\ (RMSE)\end{tabular}  \\
 \hline
    Left Clasper & 61.6 & 90.2 & 73.2 & 7.89 \\ 
    Right Clasper & 86.0 & 75.8 & 74.7 & 6.31 \\  
    Head & 83.4 & 67.9 & 74.9 & 5.32 \\ 
    Shaft & 94.8 & 71.5 & 81.5 & 8.25 \\  
    End & 90.3 & 64.0 & 74.9 & 7.70 \\\hline
    Avr & 82.8 & 72.2 & 76.1 & 7.09 \\ 
\hline
\end{tabular}
\vspace{0.2cm}
\caption{Performance of the supervised baseline on the unseen instruments with the modified post-processing which utilizes total variation measure.}
\label{table:endovis4}
\end{table}

\begin{table}[t!]
\centering
\small
\begin{tabular}{ |l|c|  }
    \hline
    \multicolumn{2}{|c|}{Experiments (Test set loss (MSE))} \\
    \hline
    SSL Algorithms & Avr. Loss \\
    \hline
    VAT ($\epsilon$ = 1) & 0.002319 \\
    VAT ($\epsilon$ = 0.1) & \textbf{0.002295} \\
    VAT ($\epsilon$ = 10)  & 0.002361 \\ \hline
    Pseudo-labeling & \textbf{0.002335} \\ \hline
    Mean teacher ($\xi$ = 1) & 0.002428 \\
    Mean teacher ($\xi$ = 0.1) & \textbf{0.002335} \\
    \hline
\end{tabular}
\vspace{0.2cm}
 \caption{An overview of the experiments and the respective test set losses. Test set loss is used to only find the better performing hyperparameters but not to compare algorithms. }
\label{table:Endovis5}
\end{table}

After establishing the right data augmentation strategies and post-processing pipeline, the unlabeled training data is utilized in semi-supervised learning context to see if further performance improvement is possible. To this end, mean teacher \cite{tarvainen2017mean}, pseudo-labeling \cite{yalniz2019billion} and VAT \cite{miyato2018virtual} algorithms are implemented and evaluated. It should be noted that for pseudo-labeling, the confidence threshold is set to be above 1000 total variation for multi-instrument cases and above 400 total variation for single instrument cases. For the mean teacher algorithm, $\alpha = 0.95$ is used for EMA. For VAT, the distance metric to compute the virtual adversarial loss is chosen to be MSE. In Table \ref{table:Endovis5}, $\xi$ represents the maximum consistency coefficient for the mean teacher algorithm and $\epsilon$ is the magnitude of the virtual adversarial noise. The maximum consistency coefficient is reached after 20k iterations for mean teacher algorithm, whereas there is no ramp-up for VAT as it was the case for the original paper as well \cite{miyato2018virtual}. The sigmoid schedule that was used by Oliver et. al. \cite{oliver2018realistic} is utilized for these experiments to determine the value of $\xi$ throughout the training. As can be seen in Table \ref{table:Endovis5}, 3 candidates with lower test set loss is selected for post-processing to enable a thorough comparison of the algorithms.

\begin{table}[t!]
\centering
\small
\resizebox{\columnwidth}{!}{
\tabcolsep 2.5pt
\begin{tabular}{|l|c|c|c|c|c|c|}
 \hline
 \multicolumn{1}{|c|}{\multirow{3}{*}{Keypoint}} & \multicolumn{6}{|c|}{F1-score / Pixel error (RMSE)} \\ \cline{2-7} 
 &  \multicolumn{3}{|c|}{Sequence 1-4 (Seen instruments)} & \multicolumn{3}{|c|}{Sequence 5-6 (Unseen instruments)} \\ \cline{2-7} 
  & VAT &  Pseudo-labeling & Mean Teacher & VAT &  Pseudo-labeling & Mean Teacher \\
 \hline
    Left Clasper    & 95.6 / 4.48 & \textbf{96.1 / 3.97 } & 95.8 / 5.34             & 74.8 / 9.43 & 77.8 / 7.57 & \textbf{80.5 / 8.87 } \\ 
    Right Clasper   & \textbf{98.8 / 2.94 } & 97.9 / 2.22 & 95.6 / 6.99             & \textbf{82.5 / 6.15 } & 68.8 / 6.39 & 71.0 / 6.28 \\  
    Head            & 99.7 / 3.43 & \textbf{100 / 3.54 } & 97.1 / 3.58              & 72.0 / 4.50 & \textbf{81.3 / 4.89 } & 71.5 / 5.15 \\ 
    Shaft           & 100 / 3.61 & \textbf{100 / 3.28 } & 96.0 / 2.90               & 81.4 / 7.98 & 87.2 / 8.24 & \textbf{88.9 / 9.48 } \\  
    End             & \textbf{100 / 5.24} & 99.9 / 6.05 & 99.1 / 5.23               & 83.0 / 8.25 & 82.3 / 8.40 & \textbf{85.6 / 8.66 } \\\hline
    Avr             & \textbf{98.8 / 3.94 } & \textbf{98.8 / 3.81 } & 96.7 / 4.81   & 78.7 / 7.26 & \textbf{79.5 / 7.10 } & \textbf{79.5 / 7.69} \\ 
 \hline
\end{tabular}}
\vspace{0.2cm}
\caption{An exhaustive comparison of different semi-supervised learning algorithms for seen and unseen instruments.}
\label{table:endovis6}
\end{table}


As can be seen on Table~\ref{table:endovis6}, semi-supervised learning methods consistently improve the ability to extrapolate to unseen geometries while maintaining high accuracy for seen instruments. It is observed that the network trained with pseudo-labeling method is more consistent across seen and unseen instruments, and therefore, this model has been selected as the final semi-supervised model. Furthermore, in Table \ref{table:endovis8}, a comparison of the supervised baseline, final semi-supervised model and state of the art in terms of f1-score and root mean squared pixel error provided. It should be noted that following Du et. al. \cite{du2018articulated}, for all the experiments a pixel threshold of 20 pixels on the original frame is used. Table~\ref{table:endovis8} shows that semi-supervised learning improves the supervised baseline in average f1-score and pixel error. Plus, the state of the art is improved in terms of average f1-score and pixel error while only using $\sim$2.1M trainable parameters which goes to show the usefulness of semi-supervised learning and the strength of the designed network architecture. The runtime of DAU-Net-64-3 is measured as 35 ms on Nvidia TITAN X GPU. 


\begin{table}[t!]
\centering
\small
\begin{tabular}{|l|c|c|c|}
 \hline
 \multicolumn{4}{|c|}{All sequences (F1-score / Pixel error (RMSE))} \\
 \hline
  & Supervised &  Pseudo-labeling & Du et. al \cite{du2018articulated}  \\
 \hline
Left Clasper & 80.6 / 6.67 & 82.8 / 6.63 & \textbf{86.4 / 5.03 } \\ 
    Right Clasper & 81.7 / 5.58 & 76.2 / 5.39 & \textbf{85.7 / 5.40 } \\  
    Head & 80.9 / 5.19 & \textbf{85.6 / 4.56 } & 76.3 / 6.55 \\ 
    Shaft & 85.2 / 7.25 & 90.1 / 7.32 & \textbf{91.0 / 8.63 } \\  
    End & 81.4 / 7.26 & \textbf{86.4 / 7.84 } & 77.3 / 9.17 \\\hline
    Avr & 82.0 / 6.39 & \textbf{84.2 / 6.35 } & 83.3 / 6.96 \\ 

 \hline
\end{tabular}
\vspace{0.2cm}
\caption{A comparison of f1-score and root mean squared pixel error for the supervised baseline, selected semi-supervised model and the state of the art.}
\label{table:endovis8}
\end{table}


\section{Conclusion}

This study encompasses an evaluation of semi-supervised learning for 2D-pose estimation for surgical instruments where data labeling is prone to human errors and requires a lot of time investment. All in all, it is observed that utilization of the attention mechanism improves the performance drastically and eliminates the need for a 2-stage pipeline that consists of detection and refinement. Furthermore, it has been shown that semi-supervised learning improves the performance for unseen instruments while maintaining high accuracy for seen ones. More specifically, it is recognized that the combination of pseudo-labeling and total variation is more consistent and easier to use, whereas VAT and mean teacher algorithms require extensive hyperparameter search and additional computational overhead during training. Furthermore, the introduced confidence measure, total variation, is shown to be very useful in many aspects. Our experiments indicate that the utilization of total variation as a post-processing step and/or as a part of pseudo-labeling algorithm can yield serious performance improvement. 



\bibliographystyle{splncs04}
\bibliography{egbib}

\newpage
\clearpage
\newpage
\section*{Supplementary Material}
\label{app:bbox}
\subsection*{Bounding Box Generation from 3-point Annotation}
\label{app:bbox}

In the context of retinal microsurgery, accurate localization of tips and shaft of the instrument is considered to be more valuable compared to the accurate localization of end joint of the instrument because these 3 joints are the closest to the retina during surgery. In other words, one can also argue that the detection and/or localization of the end joint is redundant for real world applications. Considering this argument, \cite{luca_thesis} uses the below given formulation to compute tight bounding boxes around the tips and the shaft of the instruments

\begin{align*}
    \hat{x}_{min} &= min_{x \in \mathcal{J}_{x}} (p) - \Delta \\
    \hat{x}_{max} &= max_{x \in \mathcal{J}_{x}} (p) + \Delta \\
    \hat{y}_{min} &= min_{y \in \mathcal{J}_{y}} (p) - \Delta \\
    \hat{y}_{max} &= max_{y \in \mathcal{J}_{y}} (p) + \Delta \\
\end{align*}

$\newline$ 
where a bounding box $\mathcal{B}$ is defined by the coordinates ($\hat{x}_{min}$, $\hat{y}_{min}$) and ($\hat{x}_{max}$, $\hat{y}_{max}$) which correspond to the vertices of the bounding box. For a given joint set $\mathcal{J}$, these vertices are computed by finding the minimum and the maximum over all the x and y coordinates. Furthermore, a scaling variable $\Delta$ is used to determine the width and the height of the bounding box. $\Delta$ can be computed using
\begin{align*}
    \Delta = \alpha \cdot max(max_{x \in \mathcal{J}_{x}} (p) - min_{x \in \mathcal{J}_{x}} (p), \\
    max_{y \in \mathcal{J}_{y}} (p) - min_{y \in \mathcal{J}_{y}} (p))
\end{align*}
where $\alpha$ = 1 is used for all experiments.

\vspace{0.4cm}
\subsection*{Additional Experiments}

\begin{table}[b!]
\centering
\small
\begin{tabular}{|l|c|c|}
 \hline
 \multicolumn{3}{|c|}{Precision / Recall / Pixel error (RMSE)} \\
 \hline
  & DAU-Net-32-3 & Du et. al. \cite{du2018articulated}  \\
   \hline
    Tip1 & 95.3 / 95.3 / 4.97 & \textbf{99.13 / 99.13 / 5.26 } \\ 
    Tip2 & \textbf{97.9 / 97.9 / 4.78 } & 97.58 / 97.58 / 4.61 \\  
    Shaft & \textbf{100 / 100 / 3.83 } & 94.12 / 94.12 / 4.93 \\ \hline
    Avr & \textbf{97.7 / 97.7 / 4.53 } & 96.9 / 96.9 / 4.93 \\ 
    \hline
\end{tabular}
\vspace{0.2cm}
\caption{Comparison of DAU-Net-32-3 with the state of the art for 3-point annotation on RMIT dataset.}
\label{table:comp_3_point}
\end{table}

\begin{table}[t!]
\centering
\small
\begin{tabular}{|l|c|c|c|c|}
 \hline
 \multicolumn{5}{|c|}{Sequence 1-4 (Seen instruments)} \\
 \hline
  & Precision &  Recall & F1-score & \begin{tabular}[c]{@{}l@{}}Pixel error \\ (RMSE)\end{tabular}  \\
 \hline
Left Clasper & 92.5 & 100 & 96.1 & 3.97 \\ 
    Right Clasper & 95.9 & 100 & 97.9 & 2.22 \\  
    Head & 100 & 100 & 100 & 3.54 \\ 
    Shaft & 100 & 100 & 100 & 3.28 \\  
    End & 100 & 99.7 & 99.9 & 6.05 \\\hline
    Avr & 97.7 & 99.9 & 98.8 & 3.81 \\ 

 \hline
\end{tabular}
\vspace{0.2cm}
\caption{Detailed results for the pseudo-labeling algorithm for seen instruments.}
\label{table:endovis10}
\end{table}

\begin{table}[t!]
\centering
\small
\begin{tabular}{|l|c|c|c|c|}
 \hline
 \multicolumn{5}{|c|}{Sequence 5-6 (Unseen instruments)} \\
 \hline
  & Precision &  Recall & F1-score & \begin{tabular}[c]{@{}l@{}}Pixel error \\ (RMSE)\end{tabular}  \\
 \hline
Left Clasper & 63.8 & 99.6 & 77.8 & 7.57 \\ 
    Right Clasper & 66.8 & 71.0 & 68.8 & 6.39 \\  
    Head & 86.2 & 76.9 & 81.3 & 4.89 \\ 
    Shaft & 97.3 & 79.0 & 87.2 & 8.24 \\  
    End & 88.5 & 77.0 & 82.3 & 8.40 \\\hline
    Avr & 80.5 & 80.7 & 79.5 & 7.10 \\ 

 \hline
\end{tabular}
\vspace{0.2cm}
\caption{Detailed results for the pseudo-labeling algorithm for unseen instruments.}
\vspace{8cm}
\label{table:endovis9}
\end{table}

\end{document}